\documentclass[%
 reprint,
 amsmath,amssymb,
 aps,
showkeys,
]{revtex4-2}

\usepackage{graphicx}
\usepackage{dcolumn}
\usepackage{bm}
\usepackage[hidelinks]{hyperref}

\usepackage{svg}
\usepackage{textcomp}  
\usepackage{scalerel}  
\usepackage{lipsum}
\usepackage{tikz}
\usetikzlibrary{fit,arrows,quotes,calc,automata,positioning,backgrounds,arrows.meta,bending,shapes,snakes,matrix,shadows,fadings,decorations,decorations.text,decorations.markings,shapes.geometric,shadings}
\usepackage{subcaption}
\usepackage{amsmath}
\usepackage{amsfonts}
\usepackage{amssymb}
\usepackage[OT2, T1]{fontenc}
\usepackage[english]{babel}
\usepackage{mathtools}
\usepackage{hyperref}
\usepackage{mdframed}
\usepackage{csvsimple}
\usepackage{subcaption}
\usepackage{multirow}
\usepackage{tcolorbox}
\usepackage{siunitx}
\usepackage{amssymb}

\definecolor{myRED}{HTML}{D50000}
\definecolor{myBLUE}{HTML}{00FBFF}
\definecolor{myDARKBLUE}{HTML}{007ACC}
\definecolor{myGREEN}{HTML}{00FF00}
\definecolor{myMAGENTA}{HTML}{C200D6}

\def\strong#1{\textbf{#1}}

\newcommand{\NORMAL}{\textcolor{myDARKBLUE}{\texttt{Normal}}}
\newcommand{\SUCC}{\textcolor{myRED}{{$\text{\ttfamily J}_{\text{\ttfamily SUCC}}$}}}
\newcommand{\UNSUCC}{\textcolor{myMAGENTA}{{$\text{\ttfamily J}_{\text{\ttfamily UNSUCC}}$}}}

\newcommand{\DATASET}[1]{
  \tikz[baseline={($(name.base) + (0, -1.75pt)$)}]{
      \node[shape=rectangle, inner sep=2pt, inner xsep=2pt, fill=white, rounded corners=0pt, draw=black!30] (name) {
        {\tiny \bf \textsf{\hyperref[sec:datasets]{\textcolor{black!30}{DS}}}}};
  } #1}
\newcommand{\METRIC}[1]{
  \tikz[baseline={($(name.base) + (0, -1.75pt)$)}]{
    \node[shape=rectangle, inner sep=2pt, inner xsep=2pt, fill=white, rounded corners=0pt, draw=black!30] (name) {
      {\tiny \bf \textsf{\hyperref[sec:metrics]{\textcolor{black!30}{M}}}}};
  } #1}

\usepackage{soul}
\setul{0.5ex}{0.2ex}

\usepackage{booktabs}
\usepackage{svg}

\begin{document}

\title{Mass-Scale Analysis of In-the-Wild Conversations\\ Reveals Complexity Bounds on LLM Jailbreaking}

\author{Aldan Creo}
\affiliation{Valencian Research Institute for Artificial Intelligence (VRAIN), Universitat Politècnica de València, Valencia, Spain.}
\homepage{https://acmc.fyi/}
\email{research@acmc.fyi}

\author{Raul Castro Fernandez}
\affiliation{Department of Computer Science, The University of Chicago, Chicago, USA}
\email{raulcf@uchicago.edu}

\author{Manuel Cebrian}
\affiliation{Center for Automation and Robotics, Spanish National Research Council, Madrid, Spain}
\email{manuel.cebrian@csic.es}


\begin{abstract}
    As large language models (LLMs) become increasingly deployed, understanding the complexity and evolution of jailbreaking strategies is critical for AI safety.
    We present a mass-scale empirical analysis of jailbreak complexity across over 2 million real-world conversations from diverse platforms, including dedicated jailbreaking communities and general-purpose chatbots. Using a range of complexity metrics spanning probabilistic measures, lexical diversity, compression ratios, and cognitive load indicators, we find that jailbreak attempts do not exhibit significantly higher complexity than normal conversations. This pattern holds consistently across specialized jailbreaking communities and general user populations, suggesting practical bounds on attack sophistication. Temporal analysis reveals that while user attack toxicity and complexity remains stable over time, assistant response toxicity has decreased, indicating improving safety mechanisms. The absence of power-law scaling in complexity distributions further points to natural limits on jailbreak development.
    Our findings challenge the prevailing narrative of an escalating arms race between attackers and defenders, instead suggesting that LLM safety evolution is bounded by human ingenuity constraints while defensive measures continue advancing. Our results highlight critical information hazards in academic jailbreak disclosure, as sophisticated attacks exceeding current complexity baselines could disrupt the observed equilibrium and enable widespread harm before defensive adaptation.
\end{abstract}

\maketitle

\section{Introduction}
\label{sec:introduction}

\begin{table*}
    \centering
    \caption{Dataset statistics and conversation type distributions.}
    \label{tab:dataset_and_conversation_stats}
    \begin{minipage}{0.48\textwidth}
        \centering
        \subcaption{Number of examples by dataset.}
        \label{tab:number_of_examples_by_dataset}
        \begin{tabular}{lr}
            \toprule
            Dataset                                              & \# of examples \\
            \midrule
            \DATASET{LMSYS} \cite{zheng2023lmsyschat1m}          & \num{1000000}  \\
            \DATASET{WildChat} \cite{zhao2024wildchat}           & \num{990372}   \\
            \DATASET{ShareGPT} \cite{liyucheng_sharegpt90k_2023} & \num{90665}    \\
            \DATASET{GRT2} \cite{aivillage_grt2_2024}            & \num{72050}    \\
            \DATASET{OASST2} \cite{kopf2023openassistant}        & \num{65143}    \\
            \DATASET{GRT1} \cite{ai_village_defcon31_dataset}    & \num{6399}     \\
            \midrule
            Total                                                & \num{2224629}  \\
            \bottomrule
        \end{tabular}
    \end{minipage}
    \hfill
    \begin{minipage}{0.48\textwidth}
        \centering
        \subcaption{Number of examples by conversation type, as defined in Section~\ref{sec:datasets}.}
        \label{tab:number_of_examples_by_conversation_type}
        \begin{tabular}{lr}
            \toprule
            Conversation type & \hspace{0.2cm} \# of examples \\
            \midrule
            \NORMAL{}         & \num{2114400}                 \\
            \UNSUCC{}         & \num{73941}                   \\
            \SUCC{}           & \num{36288}                   \\
            \midrule
            Total             & \num{2224629}                 \\
            \bottomrule
        \end{tabular}
    \end{minipage}
\end{table*}

In the context of a rapidly evolving landscape of large language models (LLMs), the security and safety concerns associated with these systems have become increasingly pressing.
As LLMs are deployed in real-world applications and their capabilities expand, the potential for misuse and harm has also grown.

Modern LLMs are typically aligned with human values and preferences through techniques such as reinforcement learning from human feedback (RLHF) \cite{ouyang2022training}, which aims to make these systems helpful and harmless. However, these goals may conflict with each other \cite{bai2022constitutional,lindstrom2025helpful}. If a user asks ``How can I make a bomb?'', the model has typically undergone training to prioritize answers that would satisfy the user---answering the question---but at the same time, has been aligned to refrain from providing such harmful information.

While priority is generally given to safety over user satisfaction, users have found ways to circumvent such protections. The most prominent form of misuse is \textit{jailbreaking}, which refers to the practice of manipulating LLMs to bypass their built-in safety mechanisms and generate harmful or inappropriate content \cite{zou2023universal,zhu2023autodan,wichers2024gradient}. These attacks range from manual prompt engineering to sophisticated automated approaches that use gradient-based optimization to generate adversarial suffixes \cite{zou2023universal} or interpretable prompts that maintain readability while achieving high attack success rates \cite{zhu2023autodan}. Recent comprehensive surveys have documented the rapid growth and diverse attack strategies in the red teaming field \cite{lin2025against,pathade2025red,raheja2024recent}, while systematic evaluations continue to assess the effectiveness of various attack and defense approaches \cite{shang2025evolving,panfilov2025capability}. The field has also begun to explore system-level safety considerations \cite{wang2025red} and automated red teaming using sophisticated compositional frameworks \cite{xiong2025cop}.

There has been a growing body of work that seeks to understand the mechanisms behind jailbreaks, focusing on explainability and interpretability of LLMs \cite{zhou2024alignment,han2025safeswitch,pan2025hidden,bereska2024mechanistic}. The broader AI safety community has established foundational frameworks for understanding risks \cite{hendrycks2021unsolved,hendrycks2022xrisk,cebrian2025supervisionpoliciesshapelongterm} and evaluation methodologies \cite{rahwan2019machine,grey2025safety}, while constitutional AI and other alignment approaches have emerged as promising safety paradigms \cite{bai2022constitutional}. However, while these studies provide valuable insights into the inner workings of LLMs and theoretical safety considerations, the question of how jailbreaks originate and evolve in practice remains largely unexplored. In this paper, we focus on the human behind the jailbreak---what does it take to develop a jailbreak? And what does this mean for the future of LLM safety?

We approach this question by focusing on the \textbf{complexity} of jailbreaks. There is no agreed-upon single metric to quantify complexity, as it is a multifaceted concept that can be measured in various ways \cite{meister2021language,bestgen2023measuring,lai2018discourse,altmann2015statistical,burden2024conversationalcomplexityassessingrisk}. Thus, in this paper we implement a range of distinct complexity metrics spanning probabilistic measures, lexical diversity, compression ratios, discourse patterns, cognitive load, and human-judged linguistic sophistication, and we analyze the complexity of jailbreaks in a dataset of over 2 million in-the-wild conversations with LLMs, collected from multiple sources.

We develop an evaluation pipeline (Section~\ref{sec:methods}) that allows us to analyze such complexity metrics in detail and present our findings in Section~\ref{sec:results}. Then, we discuss the implications of our findings in Section~\ref{sec:discussion}, where we extract the main conclusions of our work and outline future research directions.

\section{Methods}
\label{sec:methods}

This section describes our methodological approach to analyzing the complexity of pathways to large language model harm. We begin by detailing the construction and preprocessing of our comprehensive dataset from multiple publicly available conversation collections. We then outline the complexity metrics used to quantify the sophistication of harmful interactions, including probabilistic measures (\METRIC{log-likelihood}), lexical diversity (\METRIC{type-token ratio}), compression-based complexity (\METRIC{LZW ratios}), word frequency patterns (\METRIC{Zipf analysis}), cognitive load indicators (\METRIC{working memory demands}), discourse coherence measures, readability scores, 
and basic textual features. Finally, we present our analytical framework for identifying patterns in attack complexity and temporal evolution of evasion strategies. For clarity, we show metrics as \METRIC{metric} and datasets as \DATASET{dataset}. Where relevant, we also present numeric values in the format: ``value $\pm$ standard deviation''.

\begin{tcolorbox}[colback=myMAGENTA!10,colframe=myMAGENTA!100!black,sharp corners=south,sharp corners=north, boxrule=0.4pt]
    We define jailbreaking as \textbf{any conversation that results in the generation of harmful, toxic, or inappropriate content}, regardless of user intent. This definition differs from approaches that focus on specific prompt engineering techniques by emphasizing conversational outcomes.
\end{tcolorbox}

We focus on conversational outcomes rather than user intent because intent is difficult to measure reliably in real-world datasets, and we understand harm potential to be determined by the generated content regardless of intent. We operationalize this definition as described next.

\subsection{Datasets}
\label{sec:datasets}

\begin{figure*}
    \centering
    \includegraphics[width=\textwidth,keepaspectratio]{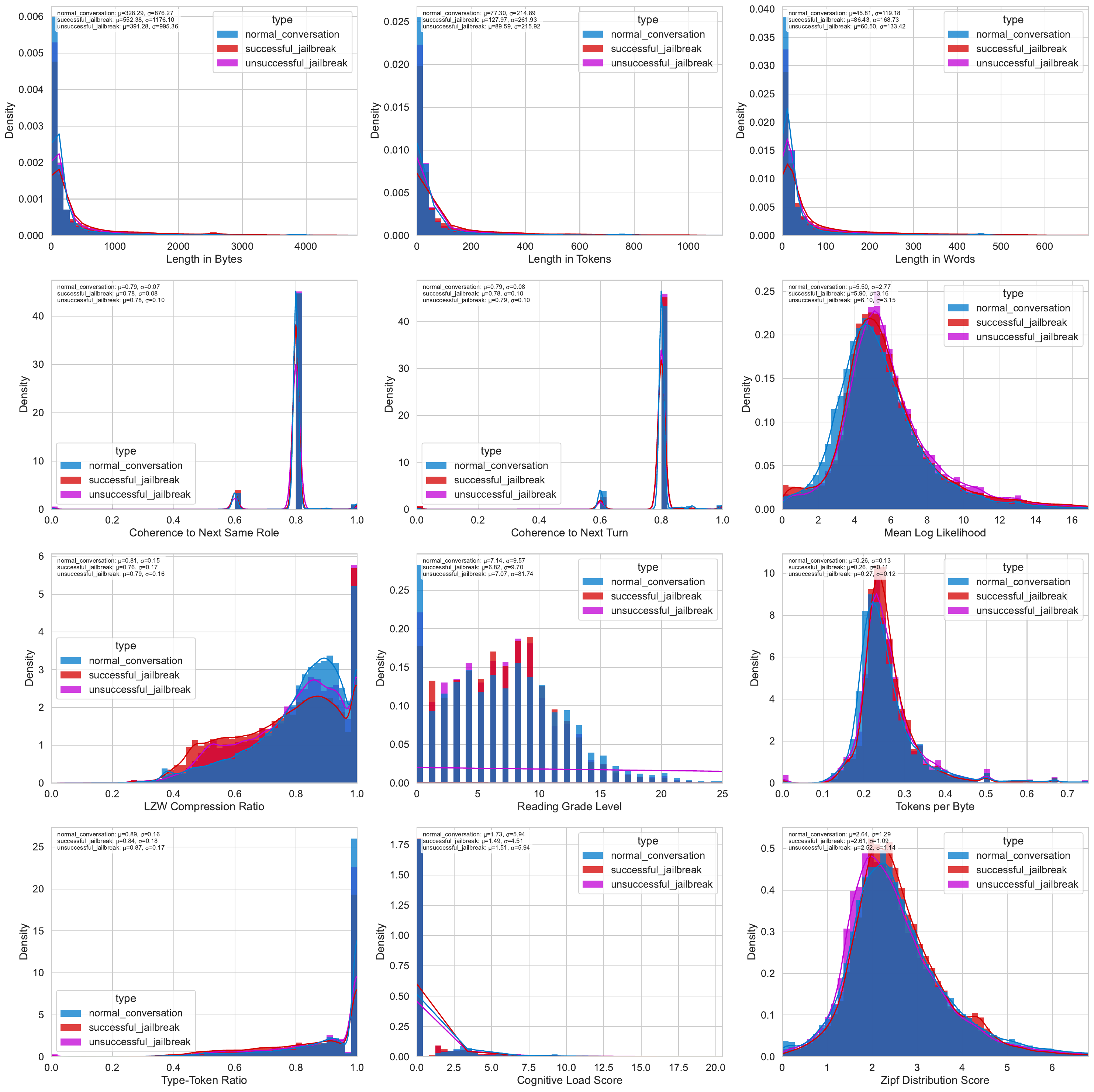}
    \caption{Complexity metric distributions by conversation type show substantial overlap between jailbreak attempts (\SUCC{}, \UNSUCC{}) and \NORMAL{} conversations.}
    \label{fig:metrics_scores_distribution_user_turns}
\end{figure*}

\begin{figure*}
    \centering
    \includegraphics[width=0.90\textwidth,keepaspectratio]{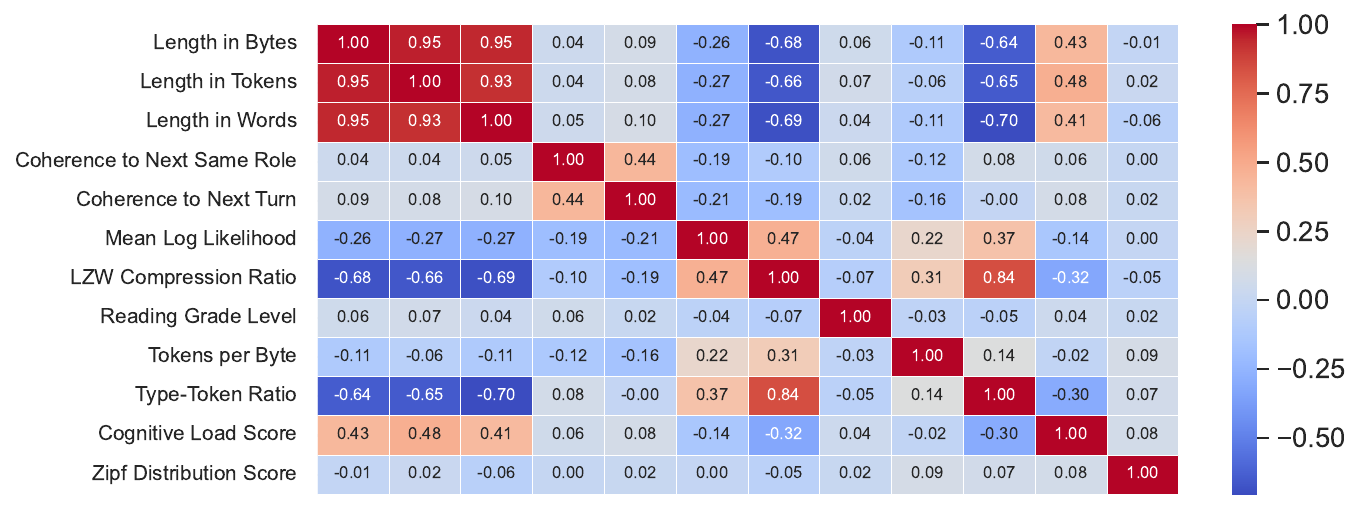}
    \caption{Correlation matrix for complexity measurements on user turns. 
    Length-based metrics are highly correlated, and correlations are moderate inside ``families'' of similar metrics, but most complexity dimensions show low correlations.
    }
    \label{fig:correlation_matrix_user_turns}
\end{figure*}

\begin{figure*}
    \centering
    \includegraphics[width=0.90\textwidth,keepaspectratio]{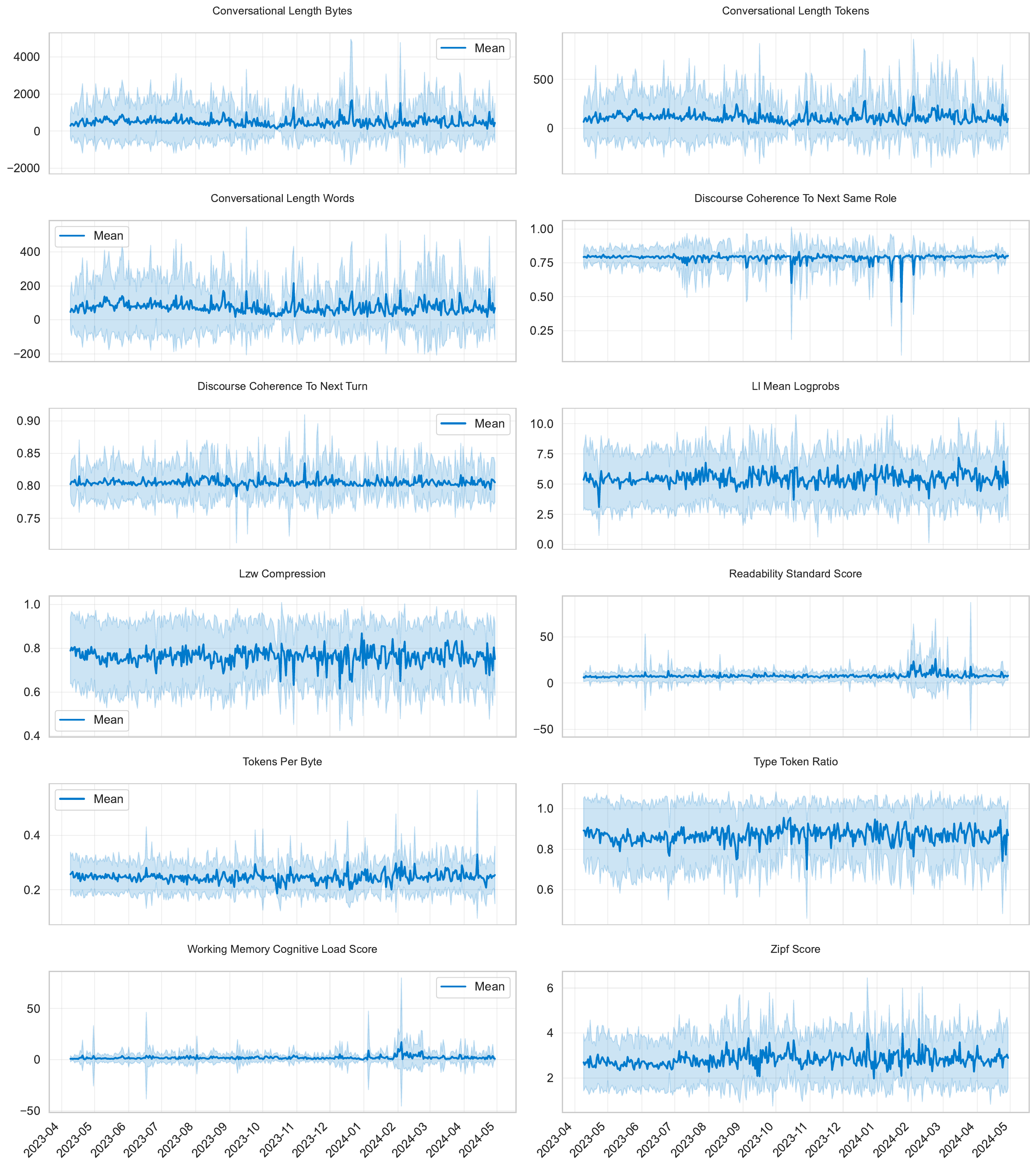}
    \caption{User interaction complexity remains generally stable over time.}
    \label{fig:temporal_complexity_trends_user}
\end{figure*}

\begin{figure*}
    \centering
    \includegraphics[width=0.85\textwidth,keepaspectratio]{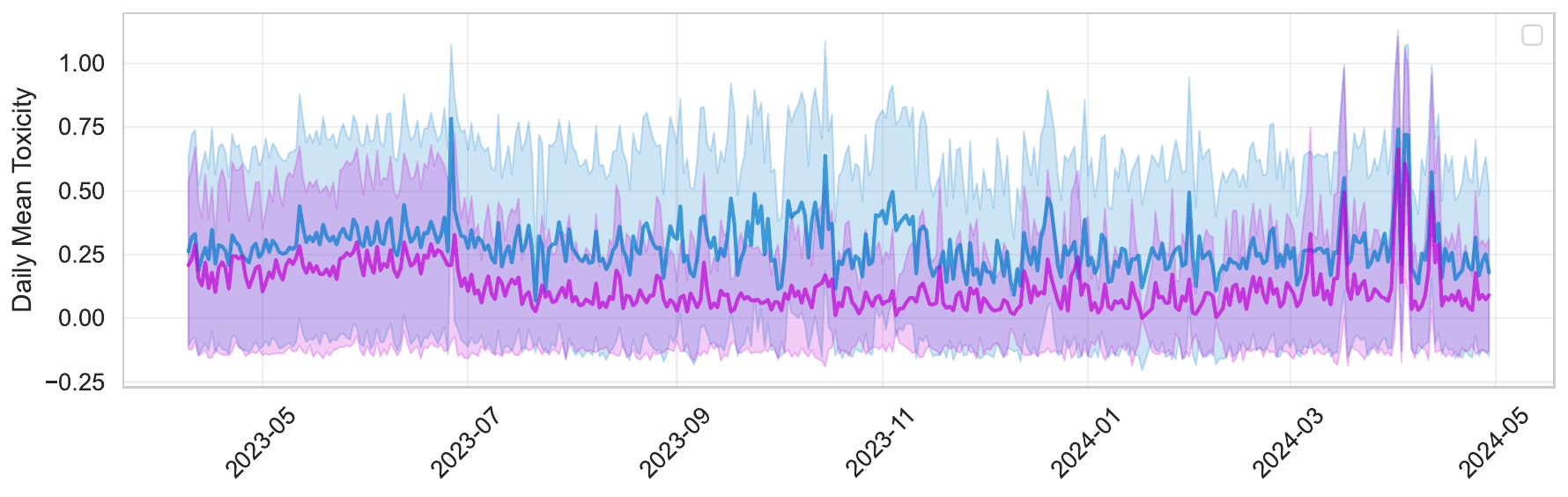}
    \caption{Assistant responses show decreasing toxicity over time while user complexity stays constant, indicating more effective safety mechanisms.}
    \label{fig:temporal_toxicity_trends}
\end{figure*}

We aggregate conversations from several publicly available datasets:

\begin{itemize}
    \item General public conversations with LLMs, representing a wide range of user interactions: \DATASET{LMSYS} \cite{zheng2023lmsyschat1m}, \DATASET{WildChat} \cite{zhao2024wildchat}, and \DATASET{OASST2} \cite{kopf2023openassistant}.
    \item Conversations from advanced communities, where users may actively seek to bypass restrictions and explore the limits of LLM capabilities: \DATASET{GRT1} \cite{ai_village_defcon31_dataset} and \DATASET{GRT2} \cite{aivillage_grt2_2024}, which are datasets collected from the AI Village community at DEFCON 31 and DEFCON 32, respectively; and \DATASET{ShareGPT} \cite{liyucheng_sharegpt90k_2023}, which is a dataset of conversations shared by users on the ShareGPT platform (discontinued), which encouraged users to share their ``wildest ChatGPT conversations''.
\end{itemize}

We present the distribution of examples across these datasets is presented in Table~\ref{tab:number_of_examples_by_dataset}. It is important to note that while these datasets include conversations from jailbreak communities, the majority of interactions are not jailbreaks. We manually inspected a sample of conversations from \DATASET{GRT1}, \DATASET{GRT2}, and \DATASET{ShareGPT} and found that most of the conversations were not jailbreak attempts, but rather users asking (e.g.) general knowledge questions or testing the system's capabilities---likely in preparation for a future jailbreak attempt.




We aggregated data from diverse sources into a unified format with conversation identifiers, message sequences, toxicity metrics, dataset metadata, and timestamps where available.
Toxicity scores were standardized using Detoxify for most datasets and OpenAI annotations for \DATASET{LMSYS}, with FastText for language identification.
Conversations were classified using toxicity threshold 0.5: \SUCC{} (successful jailbreak; both user and assistant toxic), \UNSUCC{} (user toxic, assistant not), and \NORMAL{} otherwise. Table \ref{tab:number_of_examples_by_conversation_type} shows the resulting distribution.
Data cleaning addressed dataset-specific issues including incomplete submissions and missing toxicity values, yielding our consolidated 2.2M conversation dataset.

\subsection{Subsampling}

For computational efficiency and analysis clarity, we performed stratified sampling maintaining representativeness across conversation types and datasets. We divided the target sample equally among six datasets, then subdivided among three conversation types, yielding up to \num{10000} examples per type per dataset (\num{180000} target total). Our final analysis dataset contains \num{103981} conversations.

\subsection{Metrics}
\label{sec:metrics}

We analyze conversational complexity using a range of complexity metrics:


\begin{itemize}
    \item \textbf{Length:} Message length in \METRIC{words}, \METRIC{tokens} and \METRIC{bytes} as baseline metrics.

    \item \textbf{Discourse Coherence:} Combined lexical overlap, entity coherence, and syntactic similarity between turns \cite{lai2018discourse}. We measure both \METRIC{next-turn coherence} (e.g. user-assistant) and \METRIC{next-same-role coherence} (e.g. user-user) to capture both conversational flow and intra-role dynamics.

    \item \textbf{Compression-based Complexity:} \METRIC{LZW compression} ratios to capture structural patterns and redundancy \cite{baronchelli2006artificial}.

    \item \textbf{Readability:} \METRIC{Readability} scores using a consensus combining several indices (e.g., Flesch-Kincaid Grade Level or Coleman-Liau Index) \cite{kincaid1975derivation,coleman1975computer}.

    \item \textbf{Tokenization Efficiency:} \METRIC{Tokens per byte} ratios using \texttt{Gemma-3-1B}'s tokenizer to assess encoding patterns.

    \item \textbf{Lexical Diversity:} \METRIC{Type-Token Ratio} for vocabulary richness \cite{bestgen2023measuring}.

    \item \textbf{Word Frequency Distributions:} \METRIC{Zipf's law} deviations measured via KL divergence from expected word frequency distributions \cite{altmann2015statistical}.

    \item \textbf{Cognitive Load:} \METRIC{Working memory} demands estimated through entity reference density \cite{kuribayashi2022context,shin2024evaluating}.

    \item \textbf{Probabilistic Complexity:} Mean turn negative \METRIC{log-likelihood (Ll)} computed using \texttt{Gemma-3-1B} to measure text predictability.


\end{itemize}

We compute all metrics at the turn level for both user and assistant messages, though our analysis focuses on user turns as they initiate jailbreak attempts. In all of these metrics, higher values indicate greater complexity and values are $\in \mathbb{R}$, with the exception of \METRIC{readability}, which is $\in \mathcal{Z}$.

\subsection{Statistical Analysis Considerations}
\label{sec:statistical_analysis}

To quantify differences between conversation types, we employ non-parametric statistical tests suitable for the non-normal distributions typically observed in linguistic complexity metrics. We use Kruskal-Wallis tests to assess overall differences across the three conversation types, followed by pairwise Mann-Whitney U tests for specific comparisons. Effect sizes are calculated using Cliff's Delta ($\delta \in [-1, 1]$), with extremes representing larger decreases or increases in the pairwise comparisons. $|\delta|$ < \num{0.15} is considered negligible. 
Multiple comparison corrections are applied using the False Discovery Rate (FDR) method to control for Type I errors across multiple tests.

Given our large sample sizes, we expect many statistical tests to yield significant p-values even for trivial differences. Therefore, we emphasize effect sizes over p-values when interpreting practical significance, following established guidelines that distinguish between statistical significance (detectability of differences) and practical significance (meaningfulness of differences) \cite{ellis2003practical,thompson1999improving}. With sample sizes exceeding \num{250000} user turns, our statistical power is sufficient to detect even minimal distributional differences as statistically significant, making effect size interpretation crucial for understanding practical implications.

\section{Results}\label{sec:results}

We present complexity analysis results across three dimensions: distributions, population differences, and temporal patterns. Statistical validation is provided through non-parametric tests as described in Section~\ref{sec:statistical_analysis}.


Figure~\ref{fig:metrics_scores_distribution_user_turns} shows substantial overlap across complexity measures and conversation types (\NORMAL{}, \UNSUCC{} and \SUCC{}). While Mann-Whitney U tests yield statistical significance ($p \approx$ \num{0}), effect sizes measured by Cliff's Delta remain negligible (mean $\delta$ = \num{0.01616819444(0.08456132981)}) across all 36 pairwise metric comparisons. Only three show small effects (|$\delta$| > \num{0.15}), all between \SUCC{} and \NORMAL{}: \METRIC{length in words} ($\delta$ = \num[round-mode=figures]{0.197185}), \METRIC{type-token ratio} ($\delta$ = \num[round-mode=figures]{-0.179991}), and \METRIC{LZW compression} ($\delta$ = \num[round-mode=figures]{-0.157782}). All other comparisons yield negligible effect sizes despite uniformly significant p-values. While we observe that distributions are not statistically identical based on such p-values due to our large sample size, the differences among them are generally negligible.



Statistical analysis confirms overall homogeneity in complexity patterns across diverse user populations and platforms. Cross-dataset comparisons reveal an interesting pattern: the largest effect sizes occur between datasets from the same user population—specifically between \DATASET{GRT1} and \DATASET{GRT2} (both AI Village DEFCON participants), with medium effects for \METRIC{conversational length} ($\delta$ = \num[round-mode=figures]{-0.451369}) and \METRIC{LZW compression} ($\delta$ = \num[round-mode=figures]{0.387763}). In contrast, comparisons between fundamentally different populations (specialized jailbreak communities vs general-purpose platforms) yield predominantly negligible to small effects. The vast majority (94\%) of pairwise comparisons yield negligible to small effect sizes with a mean $\delta$ = \num{-0.00319832(0.16822645644241832)}.

We also executed a correlation analysis across all metrics (Figure~\ref{fig:correlation_matrix_user_turns}), which reveals that while length-based metrics show expected high correlations, other complexity dimensions exhibit much lower correlations, with the exceptions of the two \METRIC{discourse coherence} metrics and \METRIC{mean log-likelihood}, \METRIC{LZW compression}, \METRIC{tokens per byte} and \METRIC{type-token ratio}, as well as the length-based metrics with \METRIC{cognitive load}, showing moderate correlations. In general, we observe that these correlations group into families of similar metrics, e.g., a text that is more regular (redundant), will be more compressible, both in terms of its \METRIC{LZW} compression and its \METRIC{tokens per byte} ratio, as well as having a lower \METRIC{type-token ratio} and a tendency to show a higher LLM-measured \METRIC{log-likelihood}. Additionally, we observe negative correlations between length-based metrics and both \METRIC{LZW compression} and \METRIC{type-token ratio}, which suggest that the observed longer texts are not more redundant and thus have a lower compressibility. In general, however, most complexity metric pairs are not correlated (i.e., their correlation coefficients are closer to 0).

Figures~\ref{fig:temporal_toxicity_trends} and \ref{fig:temporal_complexity_trends_user} analyze temporal toxicity and complexity patterns. Here, we focus on \DATASET{WildChat} as it provides the largest dataset with timestamps spanning a significant period. Other datasets with timestamps (\DATASET{GRT1}, \DATASET{GRT2}, \DATASET{OASST2}) are concentrated in specific times, preventing longitudinal analysis, while \DATASET{LMSYS} and \DATASET{ShareGPT} lack temporal information.

We observe that user toxicity and complexity generally remain stable over time (mean $\delta$ = \num{-0.03556713333333333(0.17607901962019645)} for toxicity; mean $\delta$ = \num{-0.025043268162393162(0.10009156382752366)} for complexity), while assistant response toxicity decreases in June 2023 ($\delta$ = \num{-0.217}). Values remain stable in other months ($|\delta| \leq$ \num{0.052}), i.e., the decrease is sustained, except for a small toxicity increase in the last month---which is not large enough to be practically significant ($\delta$ = \num{0.125}). We attribute both changes to updates in the response model; there were updates to both GPT-3.5 and GPT-4 in those periods \cite{zhao2024wildchat}. This evidence highlights the interconnection between model updates and safety mechanisms: when done correctly, updates can lead to significant improvements in safety and toxicity reduction, even if user attempts remain stable.

\begin{figure*}
    \centering
    \begin{subfigure}{0.48\textwidth}
        \centering
        \includegraphics[width=0.90\linewidth]{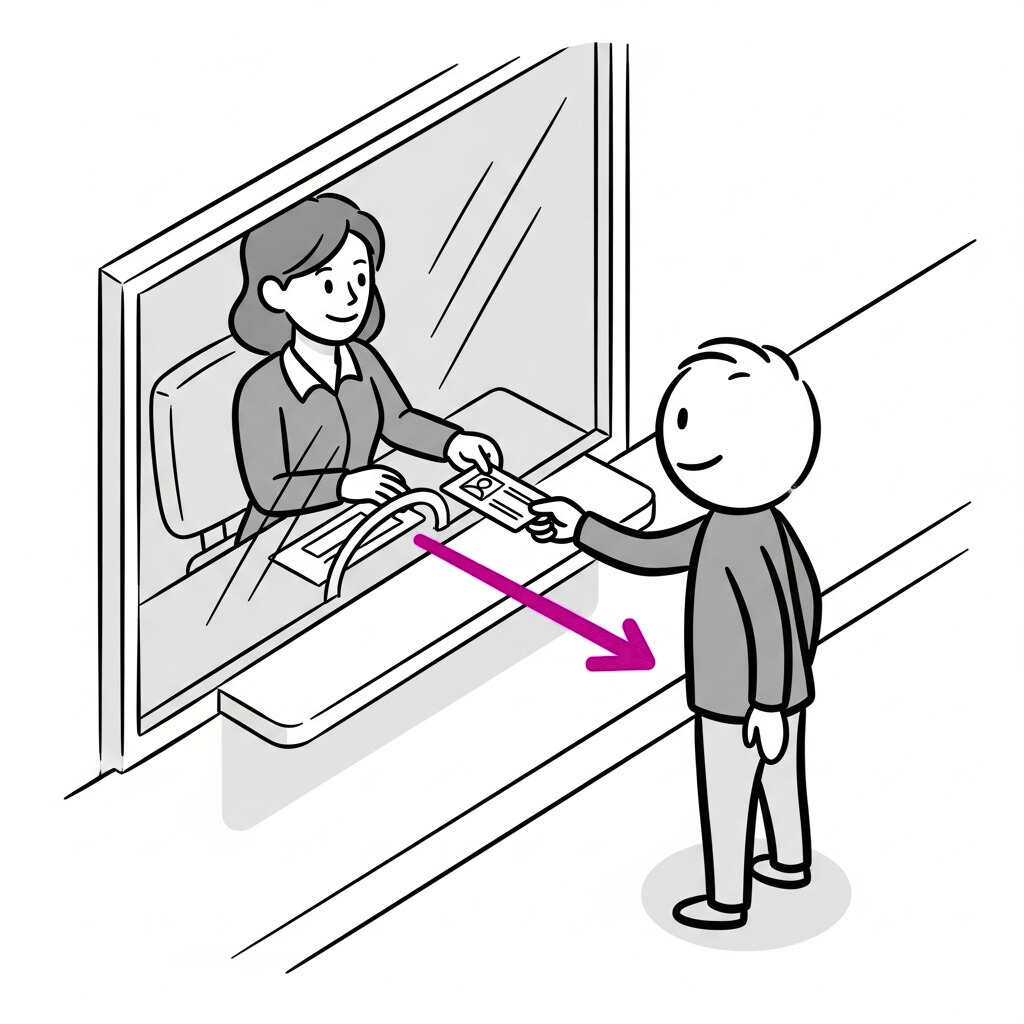}
        \caption{Using a bank as intended: simple, routine, low complexity.}
    \end{subfigure}
    \hfill
    \begin{subfigure}{0.48\textwidth}
        \centering
        \includegraphics[width=0.90\linewidth]{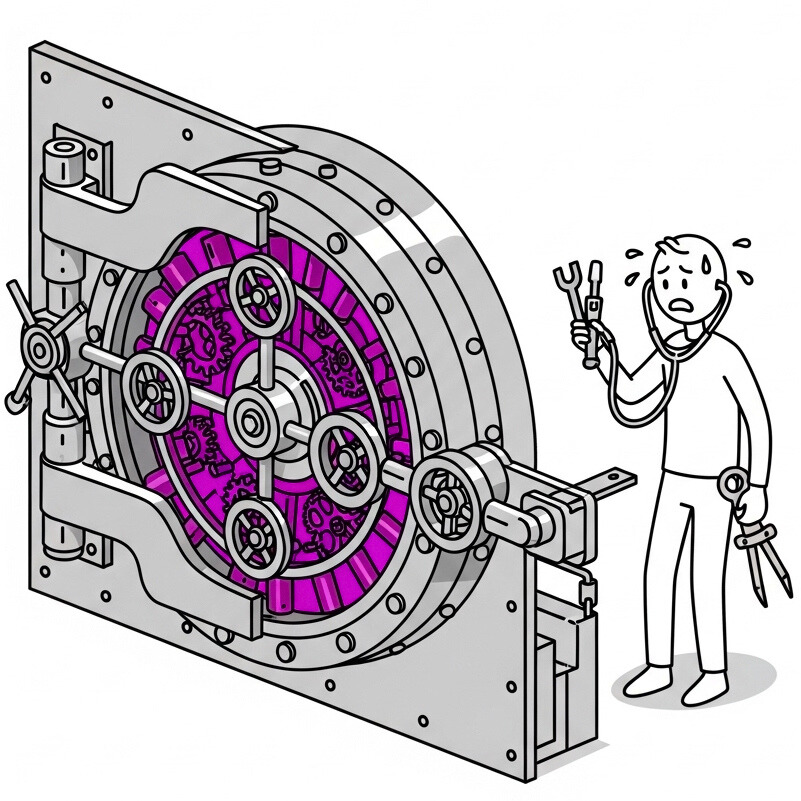}
        \caption{Trying to rob a bank: complex, risky, requires planning and expertise.}
    \end{subfigure}
    \caption{\strong{Intuition.}
        We expect breaking a system to be much more complex than using it as intended---just as robbing a bank is far more complicated than making a withdrawal at the counter. However, our data shows that \strong{real-world jailbreaks are not significantly more complex than normal conversations}. This challenges existing narratives about the complexity of jailbreaks: it appears to be bounded in practice.
    }
    \label{fig:bank_analogy}
\end{figure*}



To further investigate the distribution of the complexity of jailbreaks, we conducted a power-law analysis across all complexity metrics and conversation types. We analyzed our 12 complexity metrics across the 3 conversation types (36 total combinations), fitting power-law distributions and comparing them against exponential and lognormal alternatives using likelihood ratio tests. We define power-law behavior as p $> \num{0.05}$ for comparisons against both exponential and lognormal distributions and we find that none of the 36 combinations exhibit such behavior. This absence of scale-free behavior indicates that in-the-wild jailbreak complexity follows bounded rather than unlimited scaling patterns.


Given the large volume of data across multiple datasets, metrics, and conversation types, we provide an interactive online visualizer that allows readers to explore the results in detail (Section~\ref{sec:supplementary_materials}).

\section{Discussion}\label{sec:discussion}

As discussed in Section~\ref{sec:introduction}, the complexity of jailbreaks is a key factor in understanding their nature and impact. Our findings challenge the common assumption that jailbreaks are inherently more complex and sophisticated, akin to how robbing a bank is more complex than simply making a withdrawal (Figure~\ref{fig:bank_analogy}). Our results challenge several established narratives:

\paragraph{\strong{Multiple complexity dimensions are needed.}} Our analysis revealed that most complexity dimensions are not strongly correlated. While some correlations exist across ``families'' of metrics, there is substantial independence between dimensions. Therefore, we conclude that there is no single measure that can fully encapsulate the multifaceted nature of the complexity of jailbreaks---each captures distinct aspects of conversational complexity.

\paragraph{\strong{In-the-wild jailbreak complexity is bounded.}} Across a broad set of complexity metrics, our analysis shows that real-world jailbreaks are not significantly more complex than ordinary conversations. This holds in both specialized communities and over time, and complexity distributions do not exhibit power-law behavior. Notably, the largest complexity differences occur between datasets from the same user population rather than between jailbreak and general communities, further supporting the homogeneity of complexity patterns across user types. While more sophisticated jailbreaks could exist in controlled settings or be developed in the future, users do not naturally create them, perhaps due to linguistic or cognitive constraints \cite{bergey2024umyeahproducingpredicting}.

Irrespective of origin, this practical bound challenges the common narrative of an escalating arms race in jailbreak complexity \cite{yu2024llm,shang2025evolving}; instead, what we observe is that users are ``stuck'' at a certain complexity level, which is not significantly higher than that of normal conversations: a bounded complexity ceiling.

\paragraph{\strong{Assistants are becoming safer over time.}} New models that are better at detecting and preventing jailbreaks led to a significant decrease in assistant toxicity, even as user attempts remained stable. Combined with the bounded complexity of jailbreaks, this suggests that safety mechanisms are improving and can effectively counteract jailbreak attempts. This is a positive sign for AI safety that also challenges the assumption that jailbreaks will inevitably become more sophisticated and harder to prevent.

\subsection*{Implications for AI Safety and Research}\label{sec:implications}

The implications of these findings are substantial for AI safety and research. With a bounded complexity ceiling, the AI safety community can focus on achieving robust defensive equilibrium against human-generated jailbreaks rather than preparing for an endless escalation of attack sophistication. Efforts in this area have shown positive results in practice.

Nonetheless, such efforts may be only effective because of the observed bounds in natural human jailbreaking. Our in-the-wild analysis can only reflect the complexity of jailbreaks as they have been attempted by users during the study period. Therefore, it remains to be seen whether advanced actors like researchers could challenge these bounds by developing more sophisticated jailbreaks that laypeople replicate.

This is a fragile equilibrium. However, if we understand these risks, there is reason to be optimistic that we can design safer systems and policies, with confidence that progress in defense can outpace the risks posed by everyday users. \strong{We are not in an arms race without end, but a challenge we can meet with sustained effort and careful design.}

\section{Conclusion}\label{sec:conclusion}

This work presents a mass-scale, longitudinal measurement of jailbreak complexity in real-world LLM conversations, spanning diverse user populations and goals. Across all datasets and metrics, we consistently observe that jailbreak attempts do not display greater complexity than ordinary conversations, and that this pattern holds even in communities dedicated to jailbreaking. While it remains possible that more sophisticated jailbreaks could exist, our data suggest that, in practice, users are ``stuck'' at a certain level of complexity.

This empirical regularity points to a practical ceiling on the complexity of human-generated jailbreaks, perhaps shaped by cognitive and linguistic constraints. At the same time, we find that assistant toxicity has decreased over time, indicating that safety mechanisms are improving even as user strategies remain static. Together, these trends challenge the prevailing narrative of an escalating arms race, and instead suggest that the evolution of LLM safety is bounded by the limits of human ingenuity, with defensive progress outpacing offensive adaptation.

These findings have direct implications for the AI safety community. Since the only source of truly novel, highly complex jailbreaks may be academic research, the disclosure of such attacks carries a heightened information hazard: a breakthrough could disrupt the current equilibrium and spread rapidly. As the field advances, responsible disclosure and a nuanced understanding of complexity bounds will be essential for balancing open research with the imperative to minimize harm.

    \makeatletter
    \newcommand{\DrawLine}{%
        \begin{tikzpicture}
            \path[use as bounding box] (0,0) -- (\linewidth,0);
            \draw[color=black,dashed,dash phase=2pt,line width=0.4pt]
            (0-\kvtcb@leftlower-\kvtcb@boxsep,0)--
            (\linewidth+\kvtcb@rightlower+\kvtcb@boxsep,0);
        \end{tikzpicture}%
        \vspace{0.5em}%
    }
    \makeatother

    \vspace*{0.5cm}

        \subsection*{Acknowledgments}


        M.C. was funded by Horizon Europe Chips JU (HORIZON-JU-Chips-2024-2-RIA, NexTArc CAR), by grant PID2023-150271NB-C21 funded by MICIU/AEI/ 10.13039/501100011033 (Spanish Ministry of Science, Innovation and University, Spanish State Research Agency). This work was supported with Google.org's support through a grant to the Fundación General CSIC. Google.org had no involvement in the design, conduct, analysis, or reporting of the research.

        The authors wish to express their deep gratitude to José Hernández-Orallo and David García for their insightful comments on the first version of the manuscript, and to the Universitat Politècnica de València for computational support.


        \subsection*{Competing Interests}
        The authors have no competing interests to declare that are relevant to the content of this article.


        \subsection*{Supplementary Materials}\label{sec:supplementary_materials}
        We make our code publicly available at \url{https://github.com/ACMCMC/risky-conversations}. Our results can be found at \url{https://huggingface.co/risky-conversations} alongside an online visualizer that allows for interactive exploration of the results (\url{https://huggingface.co/spaces/risky-conversations/Visualizer}).

\bibliography{bibliography}
\end{document}